\documentclass[letterpaper]{article}
\usepackage{uai2019}
\usepackage[margin=1in]{geometry}

\usepackage{times}
\usepackage{hyperref}       
\usepackage{url}            
\usepackage{booktabs}       
\usepackage{amsfonts}       
\usepackage{nicefrac}       
\usepackage{color}
\usepackage{natbib}
\usepackage{microtype}      
\usepackage{amsmath, amsthm, amssymb, amsfonts, mathtools, graphicx, enumerate}
\usepackage{caption, subcaption}
\usepackage{enumitem}

\DeclareMathOperator*{\argmin}{arg\,min}

\newcommand{\E}{\mathbb{E}}

\setlist{nolistsep}
\newenvironment{my_enumerate}{
   \begin{enumerate}[leftmargin=*]
   \setlength{\itemsep}{0pt}
   \setlength{\parskip}{0pt}
   \setlength{\parsep}{0pt}
   \setlength{\topsep}{0pt}
   \setlength{\partopsep}{0pt}
}{
   \end{enumerate}
}

\newcount\Comments  
\definecolor{darkred}{rgb}{0.7,0,0}
\definecolor{teal}{rgb}{0.3,0.8,0.8}
\definecolor{forestgreen}{rgb}{0.13, 0.55, 0.13}
\newcommand{\kibitz}[2]{\ifnum\Comments=1{\textcolor{#1}{\textsf{\footnotesize #2}}}\fi}

\newcommand{\besa}[1]{\kibitz{forestgreen}{[BN: #1]}}

\title{Metareasoning in Modular Software Systems: On-the-Fly Configuration using Reinforcement Learning with Rich Contextual Representations}

\author{{\bf Aditya Modi$^1$, Debadeepta Dey$^2$, Alekh Agarwal$^2$, Adith Swaminathan$^2$,}\\
{\bf Besmira Nushi$^2$, Sean Andrist$^2$, Eric Horvitz$^2$}\\
$^1$ University of Michigan, Ann Arbor\\
$^2$ Microsoft Research, Redmond}

\begin{document}

\maketitle

\setlength{\abovedisplayskip}{5pt}
\setlength{\belowdisplayskip}{5pt}
\setlength{\textfloatsep}{5pt}

\begin{abstract}
Assemblies of modular subsystems are being pressed into service to perform sensing, reasoning, and decision making in high-stakes, time-critical tasks in such areas as transportation, healthcare, and industrial automation. We address the opportunity to maximize the utility of an overall computing system by employing reinforcement learning to guide the configuration of the set of interacting modules that comprise the system. The challenge of doing system-wide optimization is a combinatorial problem. Local attempts to boost the performance of a specific module by modifying its configuration often leads to losses in overall utility of the system's performance as the distribution of inputs to downstream modules changes drastically.  We present metareasoning techniques which consider a rich representation of the input, monitor the state of the entire pipeline, and adjust the configuration of modules on-the-fly so as to maximize the utility of a system's operation. We show significant improvement in both real-world and synthetic pipelines across a variety of reinforcement learning techniques.

\end{abstract}

\section{INTRODUCTION}
\label{introduction}

\begin{figure}
    \centering
    \includegraphics[width=0.9\columnwidth]{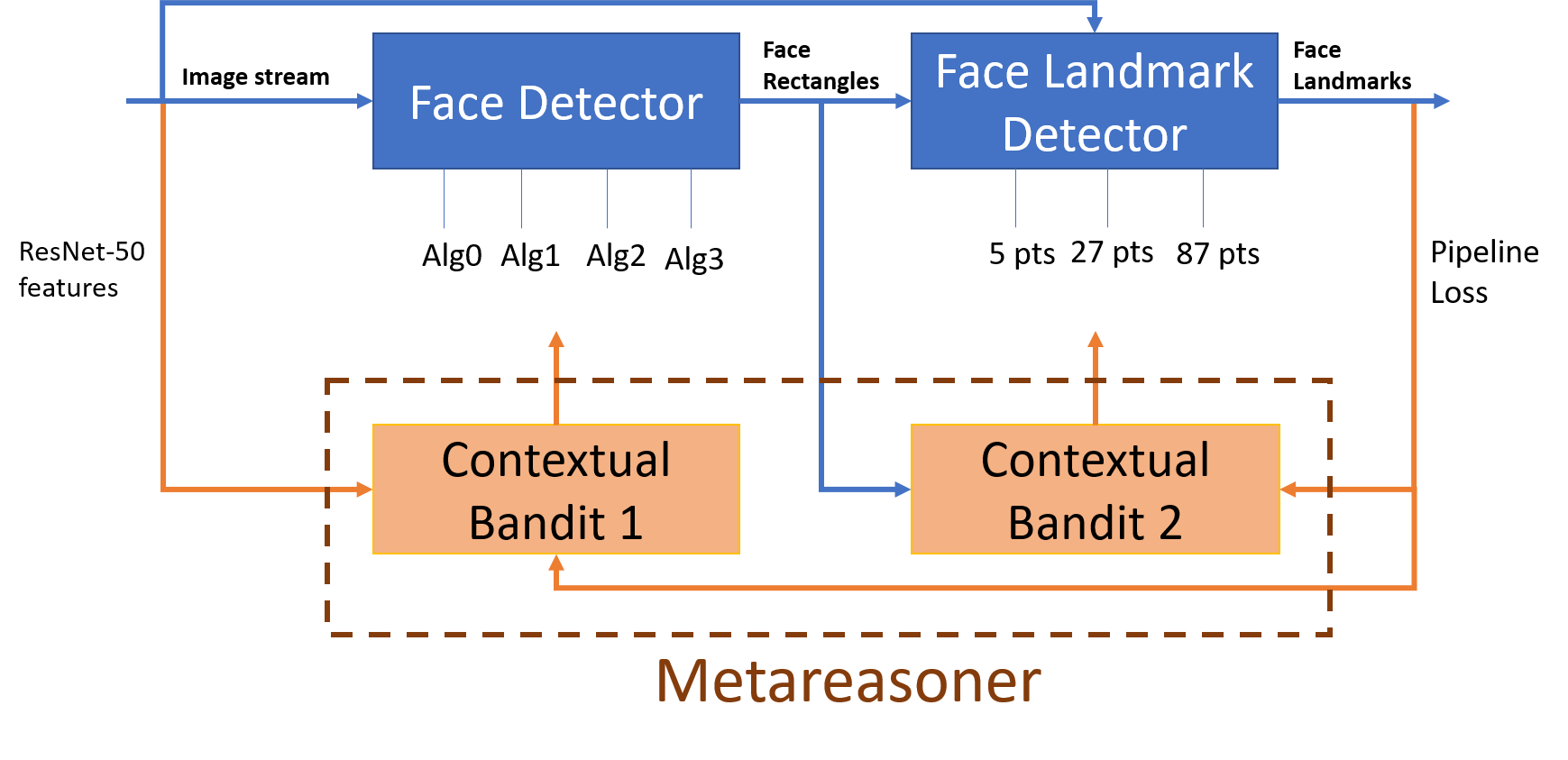}
    \caption{Face detection and landmark detection modular system. The input is an image stream to the face detection module which outputs locations of faces in the image which are then input to the face landmark detection module which outputs locations of eyes, nose, lips, brows etc on the detected face landmark modules. The metareasoning module receives the input stream of images along with intermediate outputs of the face detector to dynamically decide the configuration of the pipeline such that it optimizes the end system loss.}
    \label{fig:face_pipeline}
\end{figure}
The lives of a large segment of the world's population are greatly influenced by complex software systems, be it the software that returns search results, enables the purchase of an airplane ticket, or runs a self-driving car. Software systems are inherently modular, i.e.\ they are composed of numerous distinct modules working together. As an example, a self-driving car has modules for sensors such as cameras, lidars which poll the sensors and output sensor messages, and a mapping module that consumes sensor messages and creates a high-resolution map of the immediate environment. The output of the mapping module is then input to a planning module whose job is to create safe trajectories for the vehicle. These distinct modules often operate at different frequencies; the camera module may be producing images at  $120$Hz while the GPS module may be producing vehicle position readings at $1000$Hz. Furthermore, they may each have their own set of free parameters which are set via access of a configuration file at startup. For example, the software serving as the driver of a camera in the self-driving pipeline may have a parameter setting for the rate at which images are polled from the camera and another parameter for the resolution of the images. Similarly, the function of the mapping module may be controlled by a parameter that specifies the maximum amount of memory it is allowed to consume, leading to the continual removal of information about more distant and thus less relevant map content. 

Large software systems typically are composed of a set of distinct modular components. The operating characteristics of all of the components are usually manually configured to achieve system performance targets or constraints like accuracy and/or latency of output. Configurations of parameters may result from the tedious and long-term tuning of one parameter at a time. Once such nominal configurations have been produced, they are then held constant during system execution. The reliance on such fixed policies in a dynamic world may often be suboptimal. As an example, modules may take different amounts of time depending on the specific contents of the inputs they receive. 

As a running example, we illustrate a pipeline for extracting faces with keypoint annotations from images in Figure~\ref{fig:face_pipeline}. A natural performance metric for the pipeline might blend the prediction latency and accuracy, where the latency of a face-detection module may vary dramatically based on the number of people in the camera view. In this case, one might prefer switching to a parameter setting which allows the face detector to sacrifice some accuracy but which is much faster hence raising the overall utility of the entire pipeline. Also modules which are upstream from the face detector like the camera driver module might ideally throttle back the rate at which it is producing images since most of these images will not get processed anyways, due to a bottleneck at the face detector module.  Attempts to separately optimize distinct modules can often lead to losses in utility \cite{bradley2010learning} because of unaccounted shifts in the distribution of outputs produced by upstream modules. 

Revisiting the self-driving car example, a basic utility function is to simply navigate passengers to their destination safely and in a reasonable amount of time. Highlighting the contextuality again, the emphasis on driving time might be higher when trying to get to an important meeting or a flight than going grocery shopping. Furthermore, the utility function will typically be deeply personal to the user and has to be inferred over time. Importantly, this is a complex pipeline-level feedback which is hard to attribute to individual components.

Optimizing the configuration of large modular systems is challenging for the following reasons: 1. Changing the parameters of an upstream module can drastically change the distribution of inputs to downstream modules. Jointly choosing configuration for each module leads to a combinatorial optimization problem where the space of assignments is the cross product of the action space of the parameters of each module. 
2. Even if we solved the combinatorial optimization problem, a fixed configuration is not good across all inputs. Hence, we need to choose the configuration in an \emph{input-adaptive manner}. This decision about a particular module's parameter assignment has to be made \emph{before} input is passed through it. 3. There are challenges of credit assignment about how much each particular parameter assignment, for each module along the way, contributed to the final utility. For non-additive utility functions, this is especially challenging \cite{daume2018residual}. 4. Finally, the metareasoning process by itself should add negligible latency to the original system. If the cost of metareasoning is significant, it may be best to run the original pipeline with different configurations and select the best performing assignment.

In this work, we leverage advances in representation and reinforcement learning (RL) to develop metareasoning machinery that can optimize the configuration of modular software systems under changing inputs and compute environments. Specifically we demonstrate that by having a metareasoner continuously monitor the entire system we can switch parameters of each module on-the-fly to \emph{adapt} to changing inputs and optimize a desired objective. We also study the distinction between attainable performance between choosing the best configuration for the entire pipeline as a function of just the initial input, versus further choosing the configuration of each module based on all the preceding actions and outputs. We experiment with a synthetic pipeline meant to require adaptivity to the inputs, and we find that by doing so at each module, we improve by roughly 50\% or more over the best constant assignment, and typically by a similar margin over the choice of a configurationn just as a function of the initial input. For the face and landmark detection pipeline~\ref{fig:face_pipeline}, we use the activations of a pretrained neural network model as a contextual signal and leverage this rich representation of context in decisions about the configuration of each module \emph{before} the module operates on its inputs. We characterize the boosts in utility provided via use of this contextual information, improving $9\%$ or more across different utility functions as opposed to the best static configuration of the system. Overall, our experiments demonstrate the importance of online, adaptive configuration of each module. 

\section{RELATED WORK}

\noindent \textbf{RL to control software pipelines:} Decisions about computation under uncertainties in time and context have been described in \cite{horvitz1997perception}, which presented the use of metareasoning to guide graphics rendering under changing computational resources, considering probabilistic models of human attention so as to maximize the perceived quality of rendered content. The metareasoning guided tradeoffs in rendering quality under shifting content and time constraints in accordance with preferences encoded in a utility function. Principles for guiding proactive computation were formalized in \cite{horvitz2001principles}. \cite{Raman2013} characterize a tradeoff between computation and performance in data processing and ML pipelines, and provide a message-passing algorithm (derived by viewing pipelines as graphical models) that allows a human operator to manually navigate this tradeoff. Our work focuses on the use of metareasoning to replace the operator by seeting the best operating point for any pipeline automatically.

\cite{bradley2010learning} proposed using subgradient descent coupled with loss functions developed in imitation learning in order to jointly optimize modular robotics software pipelines which often involve planning modules, when the modules are differentiable with respect to the overall utility function. This is not suited to most real-world pipelines with modules described not with parameters but lines of code. In this work we instead develop fully general methods, which only assume the ability to evaluate the pipeline. Another form of pipeline optimization is to accordingly pick or configure the machine where each module should be executed. Methods in this ambit (\cite{mirhoseini2017device}) are complementary to this work in that optimizing the pipeline configuration per se remains a problem even with optimal device placement.

\noindent \textbf{RL in distributed system optimization}: The use of machine learning for optimizing resource allocation in distributed systems for data center and cluster management has been very well studied (\cite{lorido2014review,demirci2015survey,delimitrou2013paragon,delimitrou2014quasar}). Many of these techniques use supervised learning as well as collaborative filtering for resource assignment, which rely on the assumption of having a rich set of processes in the training data and might as a result suffer from eventual data bias for new workloads. Most recently, the use of reinforcement learning for learning policies which dynamically optimize resources such that service level agreements can be better satisfied has received a lot of attention especially with the rise of reinforcement learning with neural networks as function approximators (colloquially termed as `deep reinforcement learning' (\cite{li2017deep, arulkumaran2017brief}). Methods using model-free methods \cite{mao2016resource} based on policy-gradients \cite{williams1992simple,sutton2000policy} and Q-learning \cite{watkins1989learning,xu2012url} have shown promise as modeling such large-scale distributed systems is a challenge in itself. Similarly, RL has found impressive success in energy optimization for data centers (\cite{gao2014machine,memeti2018using}). 

\noindent \textbf{RL for scheduling in operating systems:} Even at the single machine level, RL has found promise for thread scheduling and resource allocation in operating systems. For example \cite{fedorova2007operating,hanus2013smart} use RL-based methods to learn adaptive policies which outperform the best statically optimal policy (found by solving a queuing model) as well as myopic reactive policies which greedily optimize for short term outcomes. The problem of scheduling in operating systems however differs from pipeline optimization in two fundamental ways. First, the operating system (as well as the scheduler) is oblivious to accuracy dependencies between different processes or threads. Second, due to either architectural or generality constraints, schedulers do not optimize process-level parameters but mainly focus on machine configuration.

\section{PROBLEM DEFINITION}
\subsection{FORMAL SETTING AND NOTATION}
A pipeline of $M$ modules can be viewed as a directed graph where each node $j$ is a module and an edge from $j$ to $k$ represents module $k$ consuming the output of $j$ as its input. We assume the graph does not have any cycles. Without loss of generality, let the modules be numbered according to their topological sort; i.e. $j$ refers to the index of a module in a linear ordering of the DAG. For each module $j$, we have a set of possible configurations---these are the actions that are available for the metareasoner to choose from. We denote this set by $A_j$. A module $j$ can then be viewed as a mapping from its inputs $x \in {S_j}^{in}$ to outputs $z \in {S_j}^{out}$, and each configuration $a \in A_j$ implies a different mapping.
As a running example, we will consider the face detection pipeline of Figure~\ref{fig:face_pipeline}. The pipeline contains two modules with module $1$ having $4$ choices and module $2$ having $3$ choices. The input space to the first module ${S_1}^{in}$ is the space of images (possibly in a feature space). The output space ${S_1}^{out}$ is the same as ${S_2}^{in}$ and can encode the image, the locations of faces in the image, and the latency induced by the first module. 
\besa{there is a latent assumption here which makes ${S_1}^{out}$ different from ${S_2}^{in}$, or better said ${S_2}^{in}$ is ${S_1}^{out}$ plus ${S_1}^{in}$, because the second component sees the image as well. The figure says the same thing.}
\besa{I felt like there could be more examples during the formalization in order to make the problem more concrete. For example, we could say what can be the possible parameters of a face detector. I can see that these are all described in the experiments but it might be too late?}

The quality of a pipeline's operation is measured using a loss function denoted by $L: {S_M}^{out} \mapsto \mathbb{R}$. \besa{It is unclear whether the definition of the loss function includes latency explicitly (e.g. latency + $\alpha$ * accuracy) or only implicitly because of the fact that latency can affect accuracy anyways. If I read the above definition, this is implicit. However, if I read the other sentence below (can be a complex trade-off) it seems like latency is also an input to L.} In the example pipeline of Figure~\ref{fig:face_pipeline}, the outputs from the landmark detector can be labeled by human evaluators to assess accuracy and $L$ can be a complex trade-off between the latency incurred by the overall pipeline in processing an image vs. the accuracy of the detected landmarks. If labels are not available, accuracy might be inferred from proxies such as an incorrect denial of authentication for a user based on the landmark detector output, which can be observed when the user authenticates via alternative means such as a password. Crucially, we only observe the value of this loss-function for the specific outputs $z \in {S_M}^{out}$ that the pipeline generates based on a certain configuration of actions at each module in response to an input $x$. We highlight that the loss function $L$ can be any function mapping the pipeline's final output and system state to a scalar value, such as a passenger's satisfaction with a ride in a self-driving car as discussed in Section~\ref{introduction}.
\besa{Are we also assuming that the loss function for an input $x$ does not depend on previous inputs or somehow on the current state of the pipeline? If so, we might want to remove the throttling example above regarding throwing away frames if the next component is busy anyways.}

A metareasoner can be represented as a collection of (possibly randomized) policies $\pi \coloneqq \{ \pi_1 \dots \pi_M \}$, where $\pi_j: {S_j}^{in} \mapsto \Delta(A_j)$ specifies a context-dependent configuration of the module and $\Delta(A_j)$ is the set of distributions over the action set $A_j$. We abuse the notation for ${S_j}^{in}$ here to denote any succinct representation of the preceding pipeline component's outputs, actions and system state variables which are needed to choose the appropriate action for module $j$. The pipeline receives a stream of inputs and we use $t$ to index the inputs. At time $t$, the pipeline receives an initial input $x_t^1 \in S_1^{in}$,\besa{could use $x_1^{(t)}$ so subscript is always for indexing components and upperscript is for time.} based on which an action $a_t^1 \sim \pi_1(x_t^1)$ is picked at the first module and it produces an intermediate output $z_t^1$. This induces the next input $x_t^2 \in {S_2}^{in}$ at the second module, at which point the policy $\pi_2$ is used to pick the next action and so on. At each intermediate module $j$, the input $x_t^j$ depends on the outputs of all its parents in the DAG corresponding to the pipeline and we assume that the input spaces ${S_j}^{in}$ are chosen appropriately so that a good metareasoner policy for module $j$ can solely depend on $x_t^j$ instead of having to depend explicitly on the outputs of its predecessors. Proceeding this way, the interaction between the metareasoner and the environment can be summarized as follows:
\begin{my_enumerate}
    \item $x_t^1 \in {S_1}^{in}$ is fed as input to the pipeline.
    \item metareasoner chooses actions for each module based on the output of its predecessors and induces a trajectory: $(x_t^1, a_t^1, z_t^1, \ldots,x_t^M, a_t^M, z_t^M)$; eventual output of the pipeline is $z_t^M$.
    \item Observe loss $L(z_t^M)$.
\end{my_enumerate}

Formulated this way, the task of the metareasoner can be viewed as an episodic fixed-horizon reinforcement learning problem, where the state transitions are deterministic (although the initial input can be highly stochastic, such as an image in the face detection example). Each input processed by the pipeline is an episode, the horizon is $M$, actions chosen by policies for the upstream modules affect the state distribution seen by downstream policies. The feedback is extremely sparse with the only loss being observed at the end of the pipeline. The goal of the metareasoner is to minimize its average loss: $\frac{1}{T}\sum_{t=1}^T L(z_t^M)$, and the ideal metareasoner can be described as:
\begin{equation}\argmin_{\pi_1 \dots \pi_M} \sum_{t=1}^T \E_{\pi_1,\ldots,\pi_M} \left[ L(z_t^M ) ~|~ x_t^1\right] \doteq J(\pi).\label{eqn:cost}\end{equation}
\besa{is $J(\pi)$ common notation to denote a set of policies?}

Our goal is to learn a metareasoner during the live operation of the pipeline. Since we only observe pipeline losses for the current choices of the metareasoner's policies, we must balance exploration to discover new pipeline configurations, and exploitation of previously found performant configurations. In such explore-exploit problems, we measure the average loss accumulated by our adaptive learning strategy as a benchmark; a lower loss is better. A better learning strategy will quickly identify good context-dependent configurations and hence have lower average loss as $T$ increases.
\besa{So in practice it is the case that nobody lets their pipeline learn in the wild from scratch because it is not safe, too costly etc etc. Can we make a note saying that this however does not make this work less valuable because one can imagine using this in an experimental safe environment first. In fact, it would be preferable to actually use such a technique so that you can get a reasonably good starting point for the metareasoner and then let it adapt in the real world.}

\subsection{CHALLENGES}
\label{sec:challenges}
In this section we highlight the important challenges that a metareasoner needs to address.

\noindent \emph{Combinatorial action space}: Viewing the entire pipeline as a monolothic entity, with an aim to find the best fixed assignment for each module with no input dependence, leaves the metareasoner with combinatorially many choices (every possible combination of module configurations) to consider. This can quickly become intractable even for modest pipelines (e.g. See Figure~\ref{fig:synth_plots}), despite the use of the simplest possible static policy class.

\noindent \emph{Adaptivity to inputs}: Having a static action assignment per module is overly simplistic in general and we typically need a policy for manipulating configurations that is context-sensitive. For example, in Figure~\ref{fig:fd_latencies}, we observe that the number of faces in the input image implies a fundamentally different trade-off between latency and accuracy; implying a different optimal choice for the image processing algorithm.

\noindent \emph{Credit Assignment}: Since we only observe delayed episodic reward, we do not know which module was to blame for a bad pipeline loss.

\noindent \emph{Exploration}: Pipeline optimization offers a fundamentally challenging domain for exploration. Though we employ ideas from contextual bandits here, we anticipate future directions that explore by using pipeline structure to derive better learning strategies.
\besa{the last item sounds kind of abstract. Are we trying to say that the problem setting and formulation is new?}

\section{METHODS}
The methods we outline now each address some of the challenges in Section~\ref{sec:challenges}. The simplest strategy is a non-adaptive (i.e. insensitive to the context) approach that can, however, effectively handle combinatorial actions (Section~\ref{sec:greedy_hillclimb}) to search for a locally optimal static assignment. A simple context-sensitive strategy views the pipeline optimization problem as a monolithic contextual bandit, and is vulnerable to a combinatorial scaling of complexity with pipeline size (Section~\ref{sec:global_cb}). Finally, the most sophisticated strategy we develop produces a context-adaptive policy, exploits pipeline structure to learn per-module policies and uses policy-gradient algorithms to quickly reach a locally optimal configuration policy (Section~\ref{sec:permodule_cb}).

\subsection{GREEDY HILL CLIMBING}
\label{sec:greedy_hillclimb}
The simplest (infeasible) strategy for pipeline optimization with input examples $x_1^1,\ldots,x_T^1$ is to brute-force try every possible configuration for each of the $T$ inputs and pick the configuration that accumulates the lowest loss. This strategy will identify the best non-adaptive (i.e. context-insensitive) configuration, but needs $T \cdot \prod_{j=1}^M \left| A_j \right|$ executions of the pipeline to find this configuration. Since this is typically intractable even for modest values of $T$ and $A_j$ (especially in real-time), we now describe a tractable alternative to find an approximately good configuration via random co-ordinate descent.

Rather than identifying the best configuration, suppose we aim to find a ``locally optimal'' configuration -- that is, for every module, if we held all other module configurations fixed then deviating from the current configuration can only worsen the pipeline loss.
To achieve this, we begin by randomly picking an initial configuration for each module in the pipeline. In each epoch, we first sample $K$ out of $T$ examples sampled uniformly with replacement from the dataset, where is $K$ is a hyperparameter that can be set based on the available computational budget. We then choose one of the modules $j\in\{1,2,\ldots,M\}$ uniformly at random and keep the configurations of all other modules fixed. We cycle through every possible action for that module (using, for instance, $K/\left| A_j \right|$ examples for each choice of action at this module) and pick the configuration that achieves the lowest accumulated loss. We then repeat this process until our training budget of examples is exhausted, or we cycled through every module without making a configuration change (which means we are at a local optimum). This is akin to a greedy hill-climbing strategy, and has been used in many diverse applications of combinatorial optimization as an approximate heuristic, for instance in page layout optimization~\citep{Hill2017}. More sophisticated variants of this approach can use best-arm identification techniques during each epoch, but fundamentally, this strategy finds an approximately optimal context-insensitive policy.


\subsection{GLOBAL BANDIT FROM INITIAL INPUT}
\label{sec:global_cb}
For many real-world pipelines, the modules' operating characteristics are sensitive to the initial input, meaning that a context-insensitive configuration policy can be very sub-optimal w.r.t. the pipeline loss. This motivates our approach to find a context-adaptive policy using contextual bandit (henceforth CB) algorithms.

A CB algorithm receives a context $x_t$ in each round $t$, takes an action $a \in A$ and receives a reward $r_t$. The algorithm learns a policy $\pi: x \mapsto \Delta(A)$ that is context-sensitive and adaptively trades-off exploration and exploitation to maximize $\sum_t r_t$.
In our setting $x_t$ is the input example to the pipeline, $A \doteq A_1 \times A_2 \dots \times A_M$ is the cartesian product of all module-specific configurations and the reward is simply the negative of the observed pipeline loss.

In our experiments, we use a simple CB algorithm that uses Boltzmann exploration (see e.g.~\citep{kaelbling1996reinforcement}). Concretely, the policy is represented by a parametrized scoring function $s_\theta: {S_1}^{in} \times A \mapsto \mathbb{R}$. The score for each global configuration is computed $s_\theta(x, a)$ and the policy is a softmax distribution of these scores:
\begin{equation}
\label{eq:softmax_policy}
    \pi_\theta(a \mid x) = \frac{\exp(\lambda s_\theta(x, a))}{\sum_{a'} \exp(\lambda s_\theta(x, a'))},
\end{equation}
where $\lambda > 0$ is a hyperparameter that governs the trade-off between exploration and exploitation.
The score function is typically updated using importance-weighted regression~\citep{bietti2018contextual} (henceforth \texttt{IWR}); that is, if we observe a reward $r_t$ after configuring the pipeline with action $a_t \sim \pi_\theta(a \mid x_t)$, then the score function is optimized to minimize $\frac{1}{\pi_\theta(a_t \mid x_t)}(r_t - s_\theta(x_t, a_t))^2$.

These contextual bandit algorithms can very effectively find context-sensitive policies $\pi$ and adaptively explore promising configurations. However, by viewing the entire pipeline as one monolithic object with combinatorially many actions, they cannot tractably scale to even moderate-sized pipelines.


\subsection{PER-MODULE BANDIT: USING INTERMEDIATE OBSERVATIONS}
\label{sec:permodule_cb}
The contextual bandit approach of Section~\ref{sec:global_cb} does not scale well with the size of the pipeline, but it does guarantee (under mild assumptions, like an appropriate schedule for $\lambda$, see e.g.~\citet{singh2000convergence}) that we will eventually find the best context-adaptive policy expressible by our scoring function $s_\theta$. It also does not capture the outputs of prior modules in choosing the configuration at a successor, which can be vital such as when a previous module incurs a large latency. Suppose we again relax the goal to instead find an approximately good ``locally optimal'' policy. Our key insight is to now employ a CB algorithm for each module, so that the algorithm for module $j$ only needs to reason about $A_j$ actions. Moreover, as inputs are processed by the pipeline, the metareasoner can use up-to-date information (e.g. about latencies introduced by upstream modules) as part of the context for the downstream bandit algorithm. 

One can again perform a variant of randomized co-ordinate ascent as in Section~\ref{sec:greedy_hillclimb}, holding all but one module fixed and running a CB algorithm for that module. This ensures that each bandit algorithm faces a stationary environment and can reliably identify a good context-sensitive policy quickly. However, this can be very data-inefficient; we will next sketch an actor-critic based reinforcement learning algorithm that can apply simultaneous updates to all modules.

Suppose we consider stochastic policies of the form~\eqref{eq:softmax_policy} for a module $j$, but where $x_t^j \in S_j^{in}$ and $a_t^j \in A_j$. A common approach to optimize the policy parameters $\theta$ is to directly perform stochastic gradient descent on the average loss, which results in the policy gradient algorithm. Specialized to our setting, an unbiased estimate of the gradient for the parameters $\theta_j$ of $\pi_j$, that is $\nabla_{\theta_j} J(\theta)$ (recall~\eqref{eqn:cost} is given by $L(z_t^M) \nabla_{\theta_j} \log \pi_\theta(a_t^j \mid x_t^j)$ since the loss is only incurred at the end. Typically, policy gradient techniques use an additional trained critic $C(x_t^j)$ as a baseline to reduce the variance of the gradients~\citep{konda2000actor}. We train the critic to minimize the mean squared error between the observed reward and the predicted reward, $(L(z_t^M) - C(x_t^j))^2$.

\section{EXPERIMENTS}
The algorithms discussed in the previous section are tested on two sets of pipelines: a synthetic pipeline with strong context dependence and a real-world perception pipeline. Our results show performance improvement by adaptively choosing the configuration of the pipeline. For all our experiments, we use a PyTorch based implementation \citep{paszke2017automatic} with RMSProp \citep{hinton2012neural} as the optimizer. For hyperparameter tuning, we perform a grid search over the possible choices. The common hyperparameters for both methods are:
\begin{itemize}[nosep,leftmargin=*]
    \item Learning rate  $\in \{0.0001, 0.0004, 0.001, 0.005\}$
    \item Minibatch size $\in \{5, 10, 20, 50, 100\}$
    \item $\ell_2$-weight decay factor $\in \{0.01, 0.05, 0.1, 1\}$
\end{itemize}
All our plots include 5 different runs with 5 randomly chosen random seeds with standard error regions. The specific details for each algorithm are as follows:

\noindent \textbf{Greedy hill-climbing} For finding the greedy step in each iteration, we use a minibatch of $1000$ samples per action ($K=A_j*1000$). The procedure is run until it converges to a fixed assignment. In the plots, we outline this as the non-adaptive baseline with which each method is compared. The final assignment obtained by the procedure is evaluated using Monte Carlo runs with sufficiently large number of samples from the input distribution (synthetic pipeline) or using samples present in a holdout set (face detection pipeline).

\noindent \textbf{Global contextual bandit} The policy parameters consist of a single policy that maps the input $x_t^1$ to a configuration for the entire pipeline, and policy class is a neural network with a single hidden layer. The inverse temperature coefficient for Boltzmann exploration, $\lambda$, is considered to be a hyperparameter. We use the \texttt{IWR} loss~\ with minibatches to perform updates to the policy. For hyperparameter tuning, we choose the setting with the minimum cumulative loss for the pipeline across the input stream. 

\noindent \textbf{Per-module contextual bandit} The policy function at each module is a single hidden layer neural network with a softmax layer at the end. We use the policy gradient update rule as discussed in Section~\ref{sec:permodule_cb}. The context for each module is the concatenation of the sequence of actions chosen for previous modules, current latency and the initial input to the system. Additionally, for each module, we implement a critic which predicts the final loss of the pipeline for the given context as described in Section~\ref{sec:permodule_cb}. The critic is again a single hidden layer neural network with a single output node and is trained using squared loss over the observed and predicted loss. We use the same learning rate for both networks. We use minibatches for training the networks for each module and these are concurrently updated for each minibatch. In addition, we also use entropy regularization weighted by \texttt{ent\_wt} with the policy gradient loss function \citep{haarnoja2018soft}. We tune hyperparameters using the best cumulative pipeline loss 
across the input stream.

\begin{small}
\begin{table}
    \centering
    \begin{tabular}{|c|p{4.9cm}|}
    \hline
    \textbf{Method}     &  \textbf{Hyperparameter choices}\\
    \hline
    Global CB   & $\lambda \in \{0.1, 0.3, 1, 5, 10\}$\\
    \hline
    Per-module CB   & \texttt{ent\_wt} $\in \{0.01, 0.03, 0.1, 0.3, 1\}$ \\
    \hline
    \end{tabular}
    \caption{Algorithm specific hyperparameter choices}
    \label{tab:hyper}
\end{table}
\end{small}

At a high-level, our experiments seek to uncover the importance of adaptivity to the inputs in configuring the pipeline. To capture practical trade-offs, we consider loss functions which combine the latency incurred while processing an input, 
along with the accuracy of the final prediction compared to ground truth annotations. 

\subsection{SYNTHETIC PIPELINES}
We begin with an illustrative synthetic pipeline designed to highlight: (1) benefits of adaptivity to the input over a static assignment, and (2) infeasibility of the global CB approach for even modestly long pipelines.
\begin{figure}
    \centering
    \includegraphics[width=0.95\columnwidth]{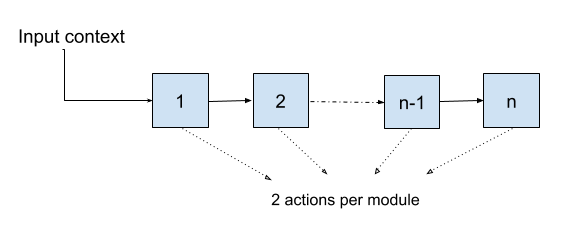}
    \caption{Synthetic pipeline}
    \label{fig:toy}
\end{figure}
The structure of the synthetic pipeline with $n$ modules is a linear chain of length $n$ as shown in Figure~\ref{fig:toy}. Each module has two possible actions: $0$ and $1$ (cheap/expensive action) which incur a latency cost of $0$ and $1$ respectively. Inputs to the pipeline consist of uniformly sampled binary strings from $\{0,1\}^n$, with the $i_{th}$ bit encoding the preferred action for module $i$. If the $i_{th}$ bit is set to $0$, both actions give an accurate output and if it is $1$, only the expensive action gives an accurate output. If we make an incorrect prediction at module $i$, then the final prediction at the end of the pipeline is always incorrect.\besa{say why the pipeline loss is constructed like this. Basically, this way you want to make sure that it is very hard to learn a policy without leveraging context} At each episode $t$, we provide an input to the pipeline by first sampling a random binary string as mentioned above, but then add uniform noise in the interval $[-0.3, 0.3]$ to each entry and this perturbed input constitutes the initial context $x_t^1$ for the pipeline. The loss function for the final output of the pipeline is
\[
\ell(a) \coloneqq \tfrac{4}{n^2} (\text{latency} - n/2)^2 + \text{error}
\]
We center the latency term at $n/2$, which is the latency of the optimal policy that routes each input perfectly to the cheapest action that makes the correct prediction for it and the normalization keeps this term in $[0,1]$. The second term measures the error in the eventual prediction, which requires each module to make an accurate prediction. The value is set to 1 for an incorrect output and 0 otherwise.
While the initial input encodes the optimal configuration,suited to global CB, there is further room to adapt. When module $i$ makes an error in prediction, then all modules $j > i$ should pick the cheap action.

\begin{figure*}[ht]
    \centering
    \begin{subfigure}[b]{0.33\linewidth}
        \centering
        \includegraphics[width=\textwidth]{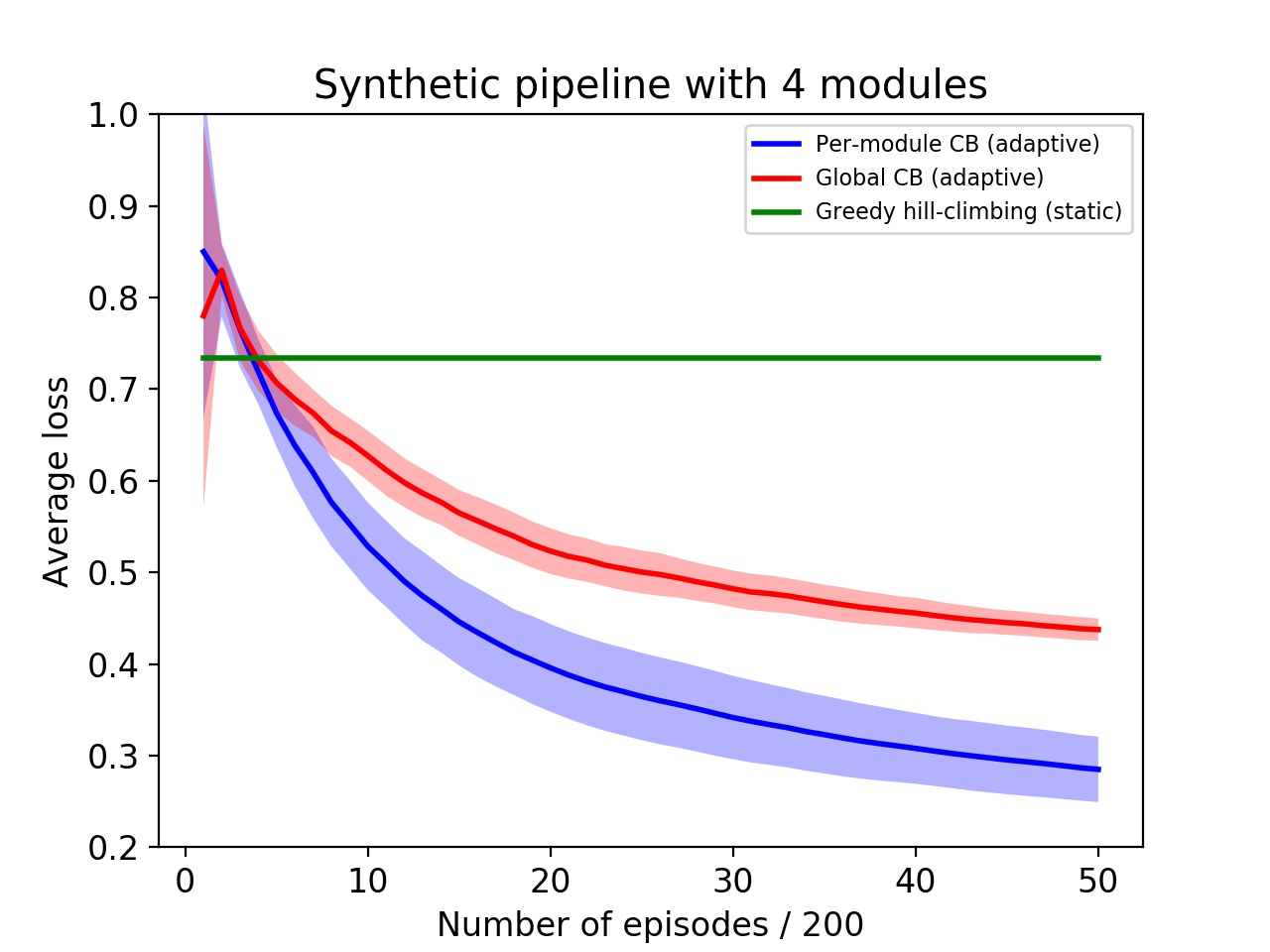}
        \label{fig:synth_4_reg}
    \end{subfigure}
    \begin{subfigure}[b]{0.33\linewidth}
        \centering
        \includegraphics[width=\textwidth]{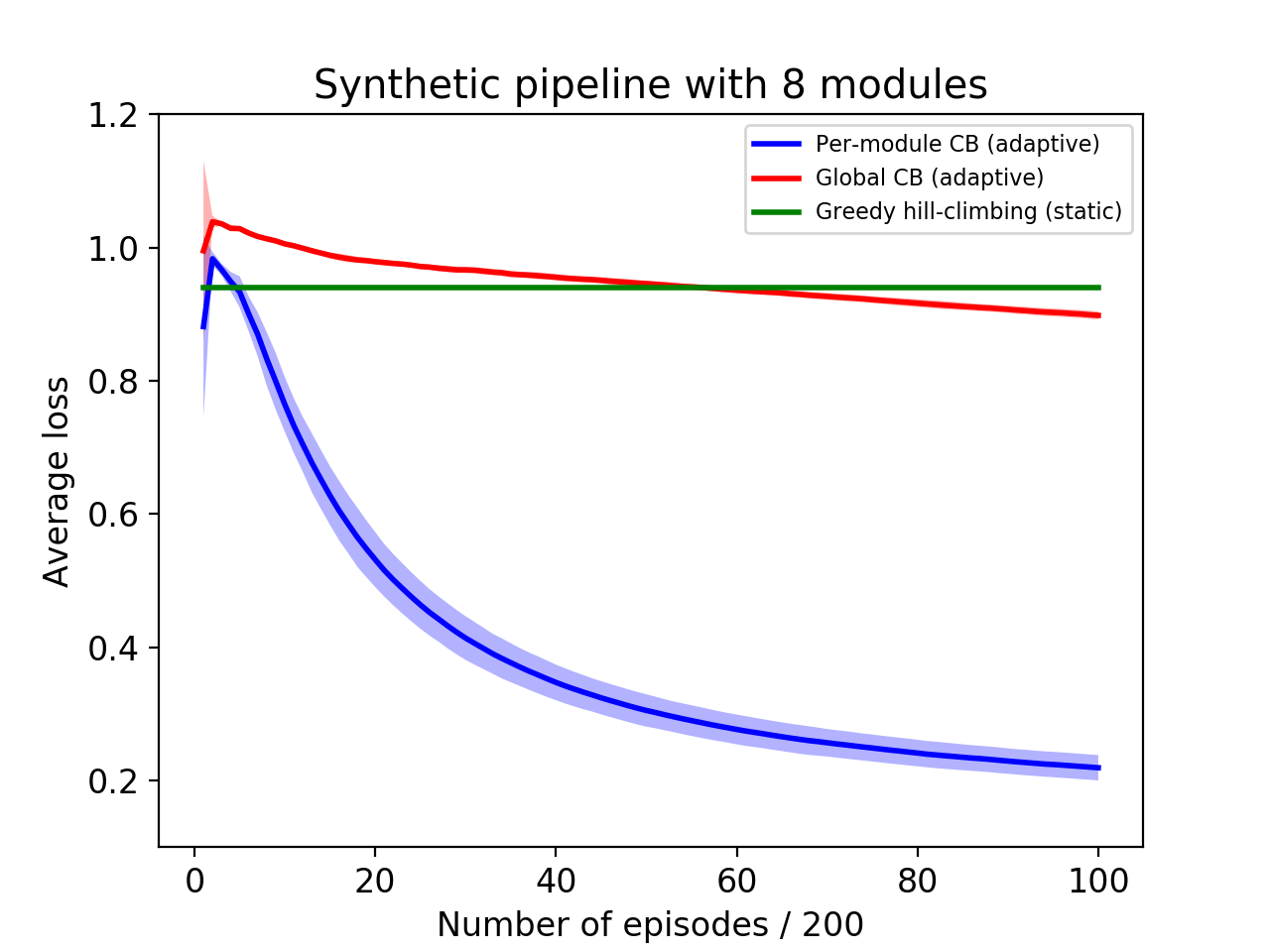}
        \label{fig:synth_8_reg}
    \end{subfigure}
    \begin{subfigure}[b]{0.33\linewidth}
        \centering
        \includegraphics[width=\textwidth]{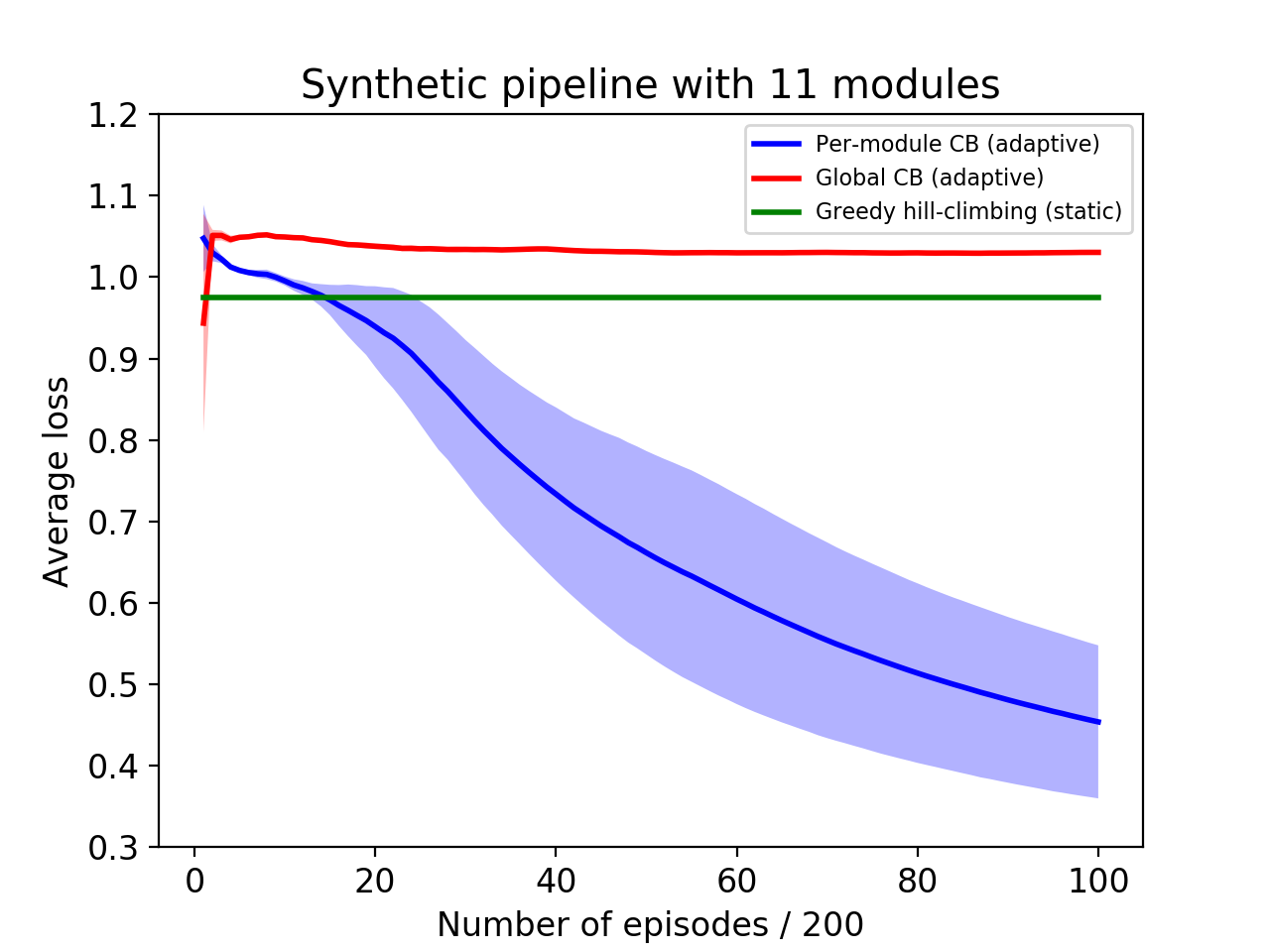}
        \label{fig:synth_11_reg}
    \end{subfigure}
    \caption{Average loss as a function of the number of examples for the synthetic pipeline. The flat line corresponds to the expected loss of the best constant assignment. The shaded region represents one standard error over $5$ runs.}
    \label{fig:synth_plots}
\end{figure*}

We show results of our algorithms for $n=4, 8$ and $11$. For static assignments, we compute both the solution of the greedy hill climbing strategy and a brute force search over all assignments, which results in similar average losses under the input distribution. The context for each module for per-module CB contains the pipeline's input, a binary string to denote upstream actions and the current latency. We use ReLU activations with the number of hidden layers for each network in our experiments as the average of input dimension and the output dimension. For instance, for global CB, the number of hidden layers for $n=4$ is $h = 10$ and for per-module CB is $h=\tfrac{d_{\text{in}}}{2} + 1$.

We show the evolution of the average loss as a function of the number of examples for different values of $n$ in Figure~\ref{fig:synth_plots}. Our results show significant gains for being adaptive over the constant assignment baseline in all the plots. For $n=4$, the total number of assignments is $16$ and it can be clearly seen that global CB is effective when compared to the per-module counterpart. However, global CB is slower in convergence than per-module CB. For $n=8$, the difference between the two is more pronounced as the per-module CB method converges rapidly. For $n=11$, the total number of assignments for the pipeline is $2048$ and global CB completely fails to learn a better adaptive policy. The per-module CB has a slower convergence in this harder case, but still improves upon the best constant assignment extremely quickly.

\subsection{FACE AND LANDMARK DETECTION}
\begin{figure*}[!hbt]
    \centering
    \begin{subfigure}[b]{0.24\linewidth}
        \centering
        \includegraphics[width=\textwidth]{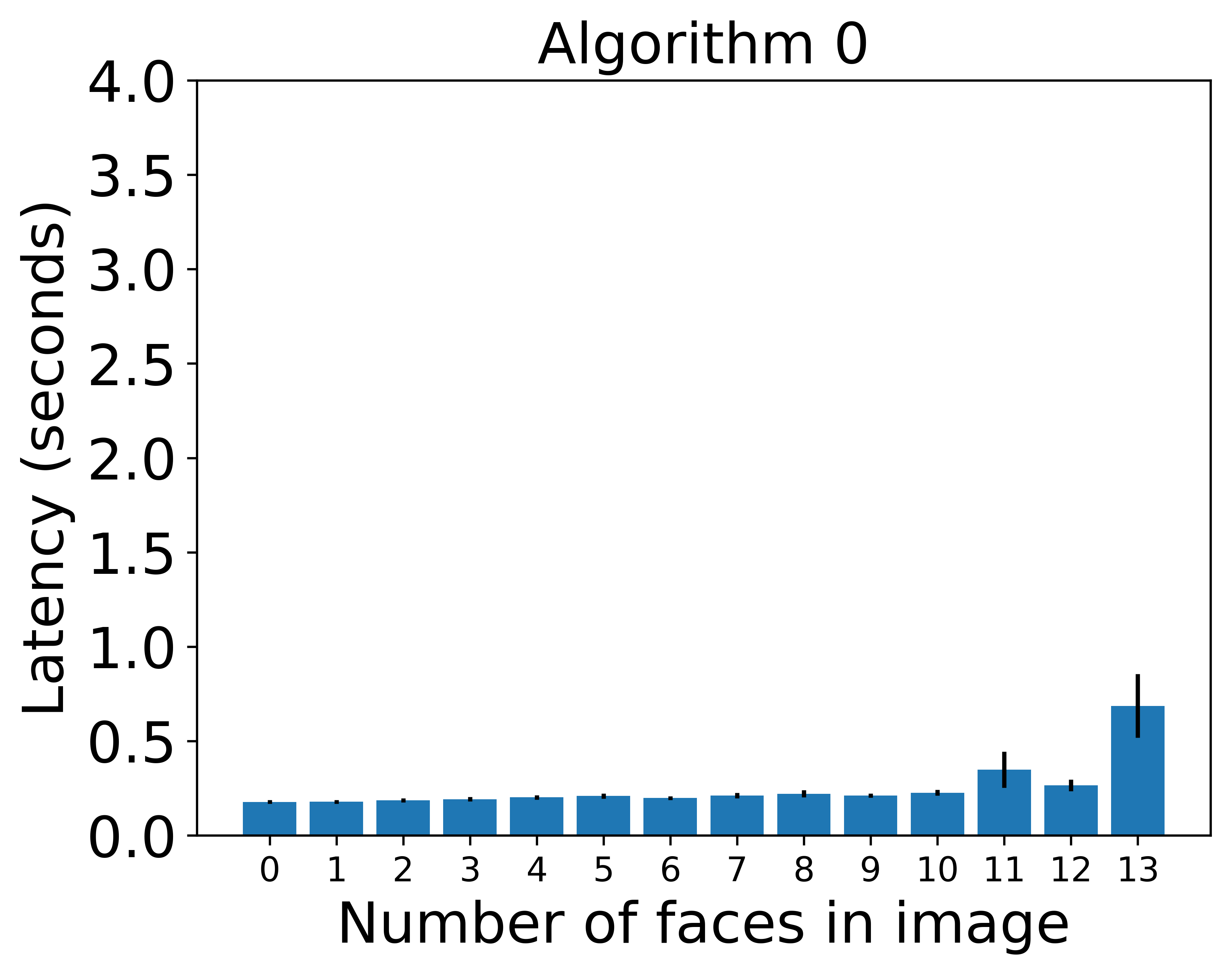}
        \label{fig:fd_alg_0}
    \end{subfigure}
    \begin{subfigure}[b]{0.24\linewidth}
        \centering
        \includegraphics[width=\textwidth]{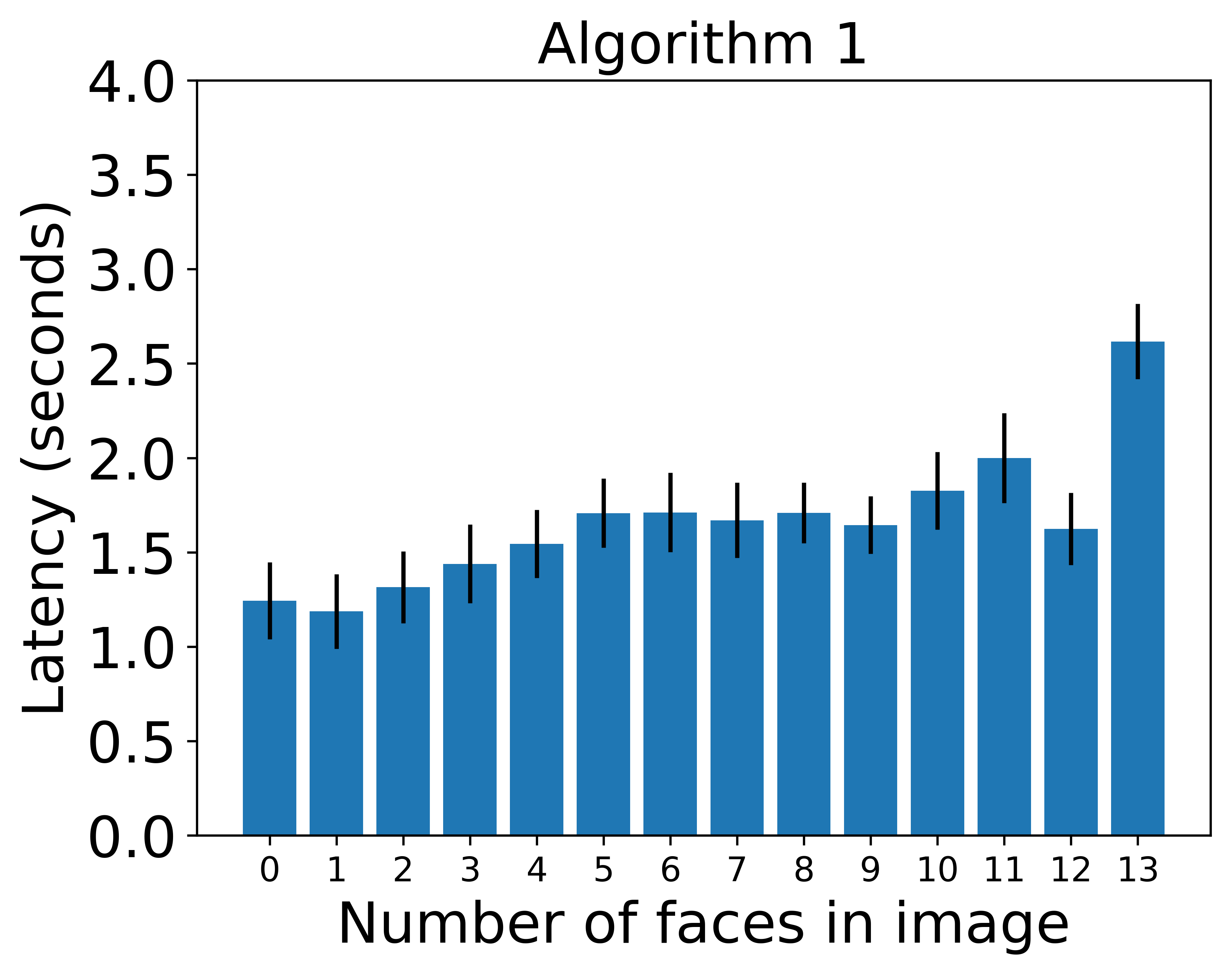}
        \label{fig:fd_alg_1}
    \end{subfigure}
    \begin{subfigure}[b]{0.24\linewidth}
        \centering
        \includegraphics[width=\textwidth]{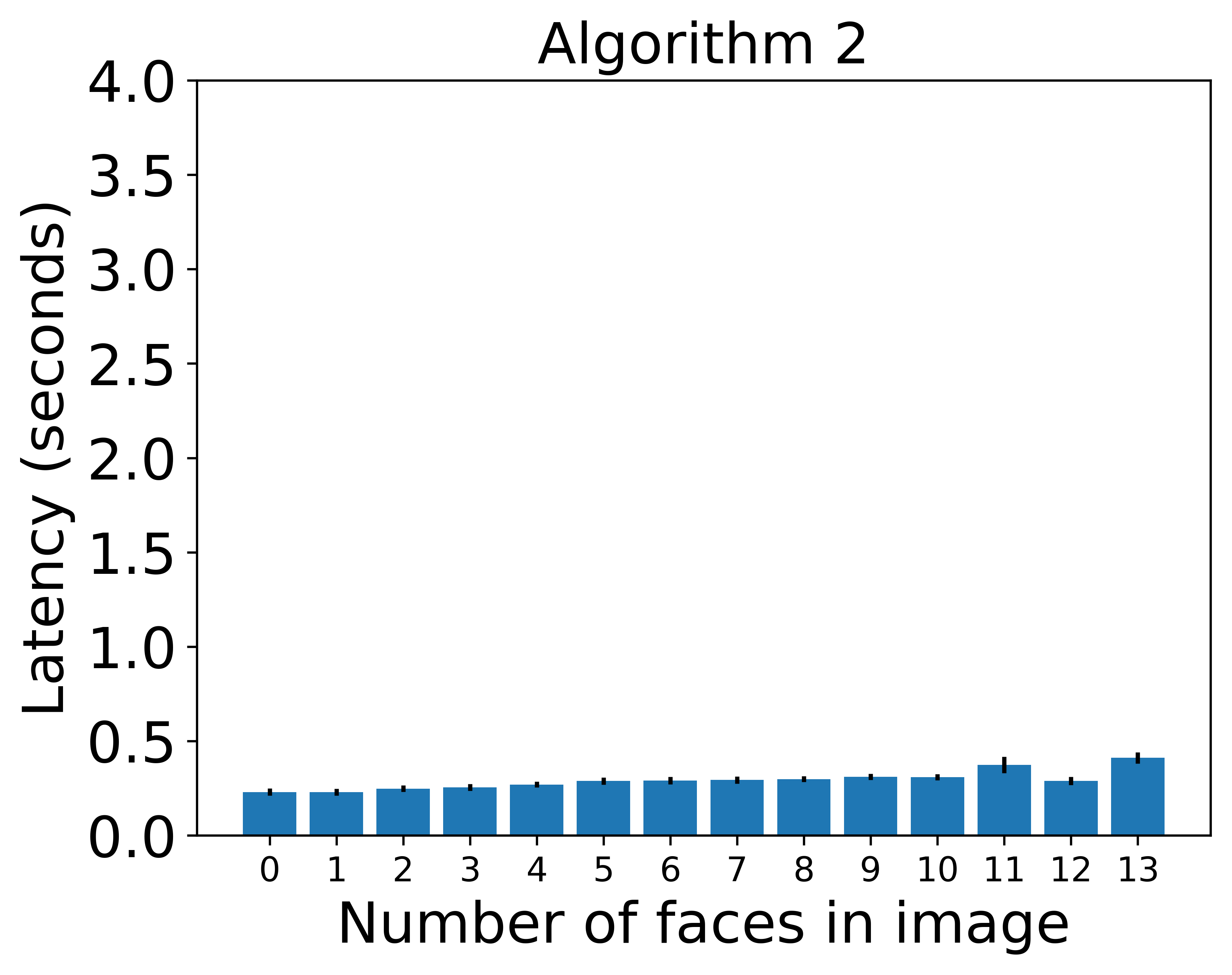}
        \label{fig:fd_alg_2}
    \end{subfigure}
    \begin{subfigure}[b]{0.24\linewidth}
        \centering
        \includegraphics[width=\textwidth]{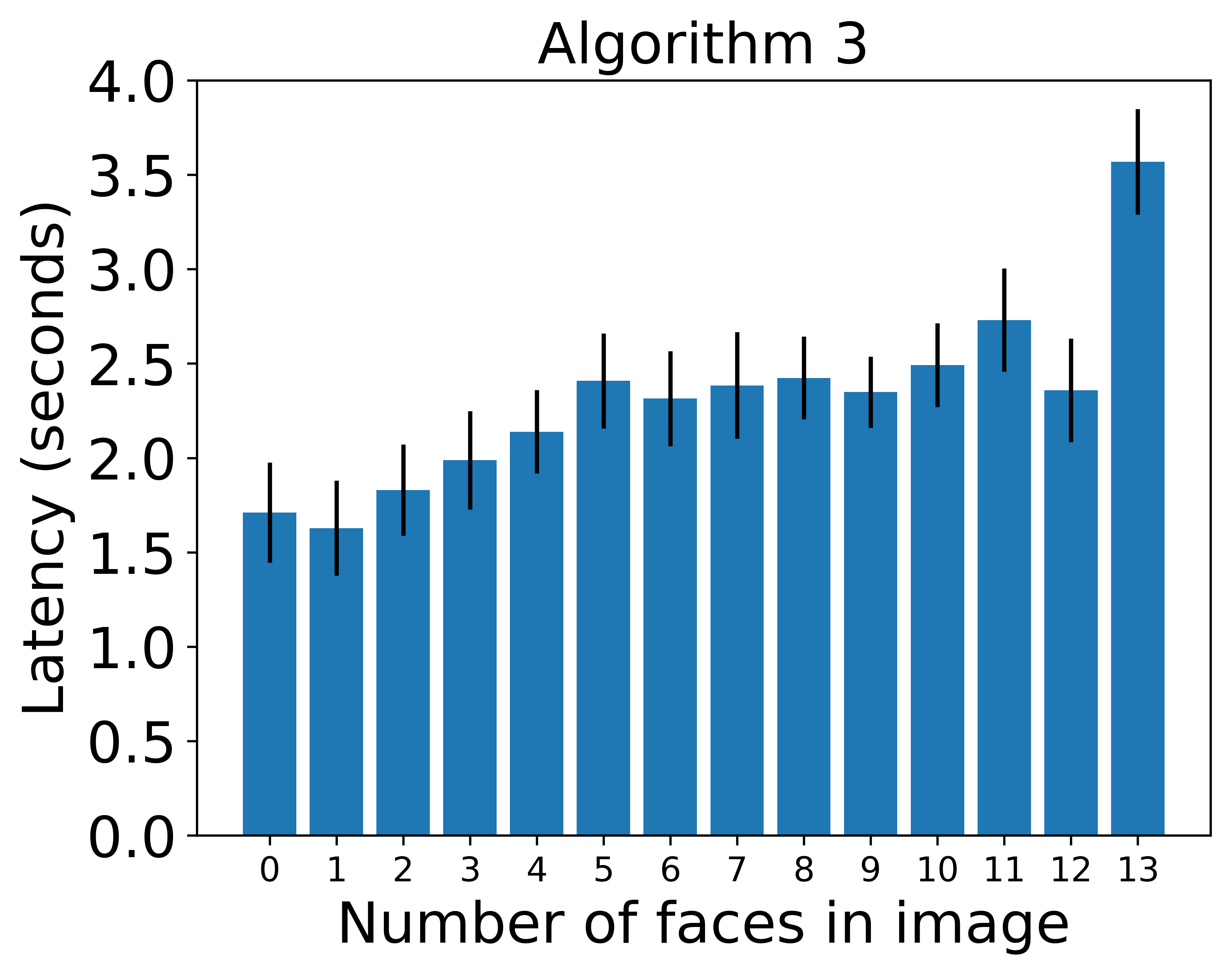}
        \label{fig:fd_alg_3}
    \end{subfigure}
    \caption{Face detection algorithm choices vs. latency in seconds as a function of true number of faces present in the image. Algorithm 0 and 2 are much faster than Algorithm 1 and 3. All algorithms exhibit increasing latencies as the number of faces goes up in the image.}
    \label{fig:fd_latencies}
\end{figure*}

\noindent \textbf{Pipeline and dataset:} We use a two-module production-grade real-world perception pipeline service to empirically study the efficacy of our proposed methods (Figure \ref{fig:face_pipeline}).
The first module is a face detection module which takes as input an image stream and outputs the location of faces present in the image as a list of  bounding box rectangles. This module has four different algorithms for detecting faces. The exact details of the algorithms are proprietary and hence we only have black-box access to them. We benchmarked the latency and accuracy of the algorithms on 2689 images from the validation set of the 2017 keypoint detection task of the open source COCO dataset \citep{lin2014microsoft}. COCO has ground truth annotations of up to 17 visible keypoints per person in an image. We notice that not only do each of the algorithm choices have large variation in latency and accuracy on average when compared to each other, more crucially their latencies and accuracies vary drastically with the number of true faces present in the incoming images, i.e.\ they are \emph{context dependent}. Specifically, we observe that latency drastically increases with the number of faces present in the image. Figure \ref{fig:fd_latencies} shows the latencies of all four detection algorithm choices vs. number of true faces present in the image. Note that different algorithms have different latencies \emph{on average} with Algorithm 0 being the fastest ($\sim 0.2$ seconds) and Algorithm 3 the slowest ($\sim 2.5$ seconds).

\begin{figure}[t]
    \centering
    \begin{subfigure}[b]{0.47\linewidth}
        \centering
        \includegraphics[width=\textwidth]{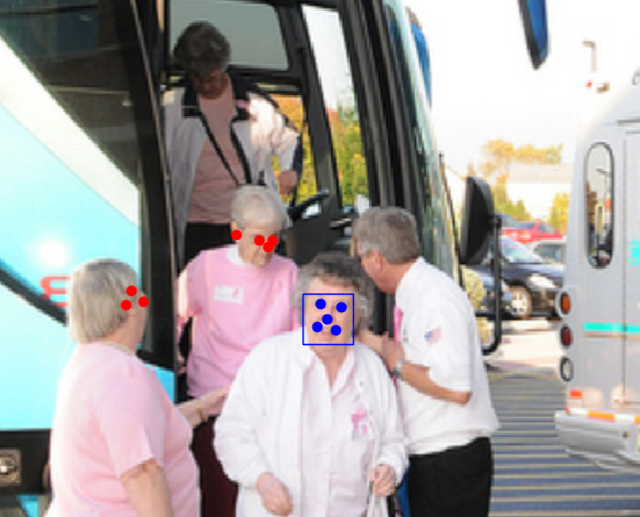}
        \label{fig:1-1}
    \end{subfigure}
    \begin{subfigure}[b]{0.47\linewidth}
        \centering
        \includegraphics[width=\textwidth]{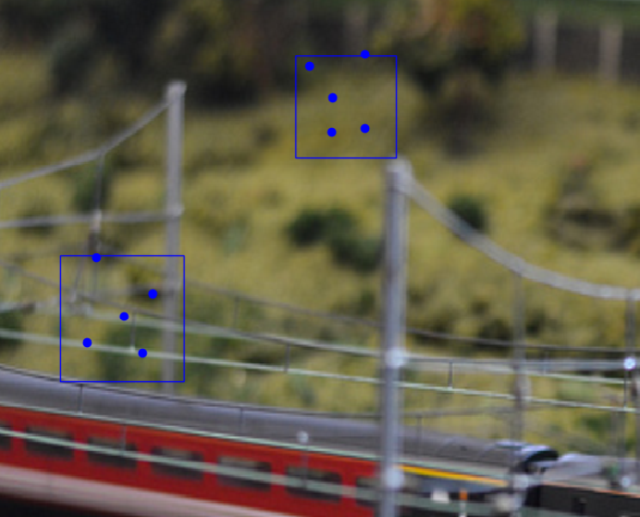}
        \label{fig:1-2}
    \end{subfigure}
    \caption{Example face and landmark detections from COCO validation set. (Left) Face detected (blue rectangle) and landmarks detected within the face (blue dots). The red dots represent groundtruth face landmarks not detected. 
    (Right) False face detections (blue rectangles) and wrong landmarks within the rectangles.}
    \label{fig:example_detections}
\end{figure}

The second module is a face landmark detection module which takes as input the original image and the predicted face rectangles output by the face detection module and computes the location of landmarks on the face like nose, eyes, ears, mouth etc. There are three different landmark detector algorithm choices: 5 points, 27 points or 87 points landmark detector. Again we observe in our benchmarking that the landmark detector which outputs 87 points takes the most time at $0.25$ ms per image on average vs. $0.17$ ms per image for the 27 points algorithm and $0.08$ ms per image for the 5 points one. Since the landmark detectors are applied on each face rectangle detected by the face detector, the computational time required goes up proportional to the number of faces. Figure \ref{fig:example_detections} shows example face detections and landmarks detected on images from the validation sets of the COCO dataset.

\noindent \textbf{Accuracy calculation}: For evaluating when a prediction by the face detection module is a true/false positive/negative, we closely follow the scheme laid out in the COCO Keypoints evaluation page \cite{lin2014microsoft}. Specifically a rectangle location on the image is considered a true positive if it is within $30$ pixels of a ground truth face annotation which is quite conservative as the images we use are all resized to constant size of $1280 \textrm{(W)} \times 960 \textrm{(H)}$ pixels. Otherwise, it is marked as a false positive. Ground truth faces which are not ``covered'' by any of the predicted faces cause an entry in the false negative count. If an image contains no faces and the face detection module also predicts no faces then we count such scenarios as true negatives.

Similarly, for the face landmark module, we mark a prediction as a true positive if it is within $5$ pixels of the ground truth landmark annotation, else a false positive. All landmarks not ``covered'' by any of the predicted faces are counted as false negatives. Note since the COCO keypoint annotations include only $17$ keypoint annotations on the entire human body including only $5$ face landmarks we don't penalize predictions of the $27$ or $87$ landmark detection algorithms which are not within threshold distance of any ground truth landmark as that unfairly counts as false positives (due to lack of ground truth annotations)\footnote{Since we find an optimal matching between predicted and true keypoints, each false negative also results in a false negative}.
\begin{figure*}[t]
    \centering
    \begin{subfigure}[b]{0.33\linewidth}
        \centering
        \includegraphics[width=\textwidth]{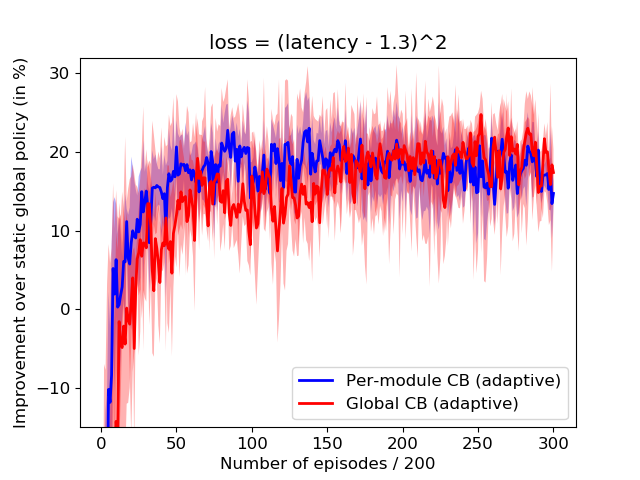}
        \label{fig:lat_1.3_reg}
    \end{subfigure}
    \begin{subfigure}[b]{0.33\linewidth}
        \centering
        \includegraphics[width=\textwidth]{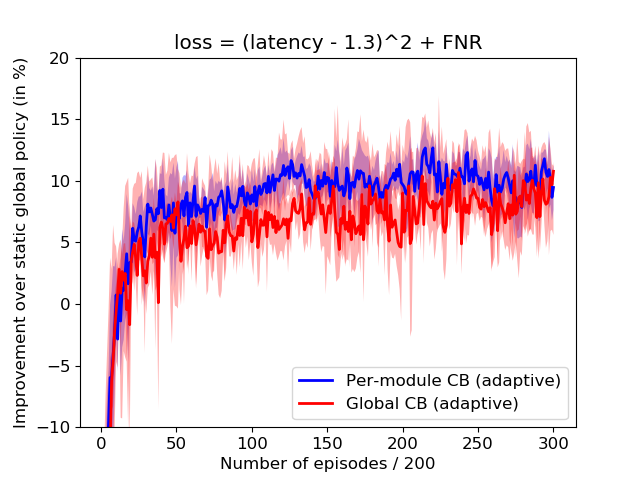}
        \label{fig:comb1_1.3_reg}
    \end{subfigure}
    \begin{subfigure}[b]{0.33\linewidth}
        \centering
        \includegraphics[width=\textwidth]{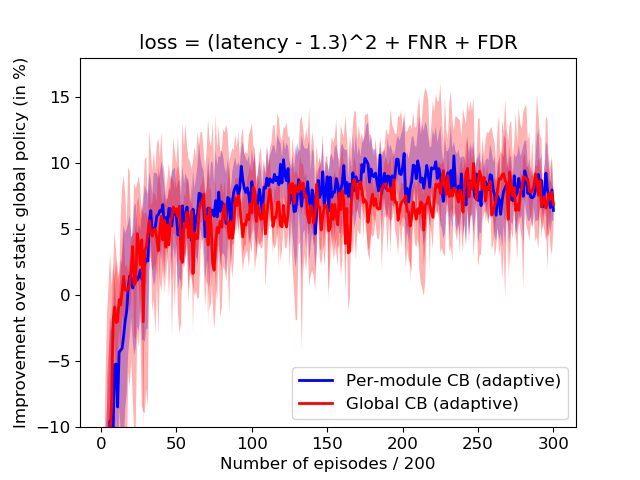}
        \label{fig:comb2_1.3_reg}
    \end{subfigure}
    \begin{subfigure}[b]{0.33\linewidth}
        \centering
        \includegraphics[width=\textwidth]{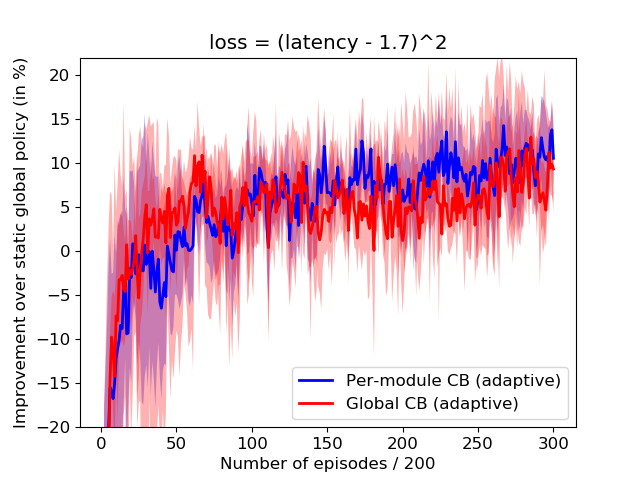}
        \label{fig:lat_1.3_test}
    \end{subfigure}
    \begin{subfigure}[b]{0.33\linewidth}
        \centering
        \includegraphics[width=\textwidth]{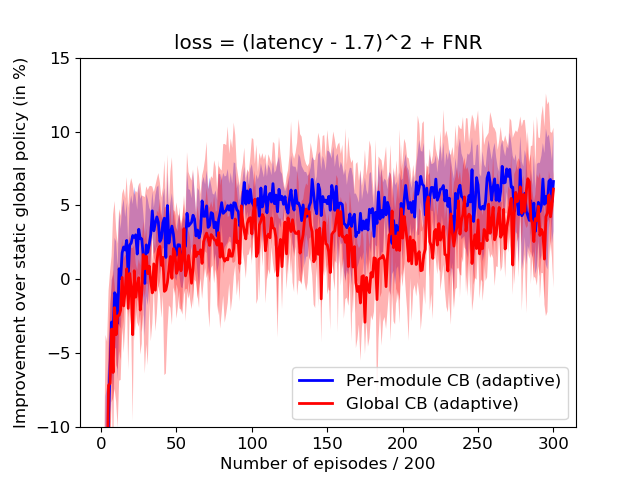}
        \label{fig:comb1_1.7_test}
    \end{subfigure}
    \begin{subfigure}[b]{0.33\linewidth}
        \centering
        \includegraphics[width=\textwidth]{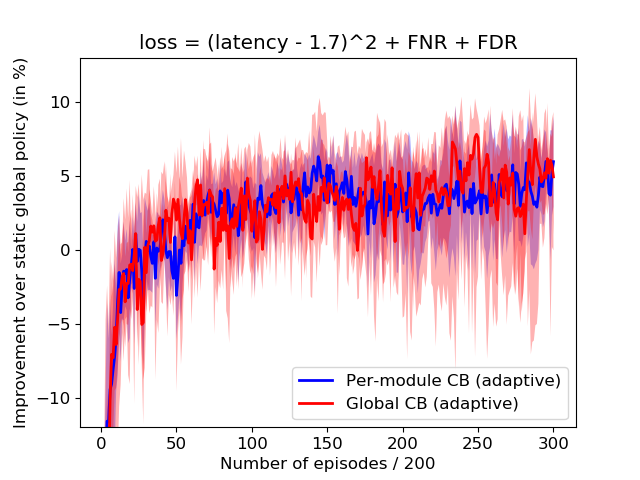}
        \label{fig:comb2_1.7_test}
    \end{subfigure}
    \caption{Test performance curves for the Face Detection and Landmark pipeline with $t_0=1.3$ and $t_0 = 1.7$. The $Y$-axis is the performance percentage improvement over static global policy after every 200 episodes of learning on held-out examples. The plots use a latency-based loss (left), latency and false negative rate (middle) and latency, false negative rate and false discovery rate (right). The adaptive approaches significantly improve over the best fixed configuration in all the cases. Shading represents standard error across 5 runs. \besa{this is the only place when we say face-sdk}}
    \label{fig:faceSDK-results}
\end{figure*}

\begin{figure}[t]
    \centering
    \begin{subfigure}[b]{0.49\linewidth}
        \centering
        \includegraphics[width=\textwidth]{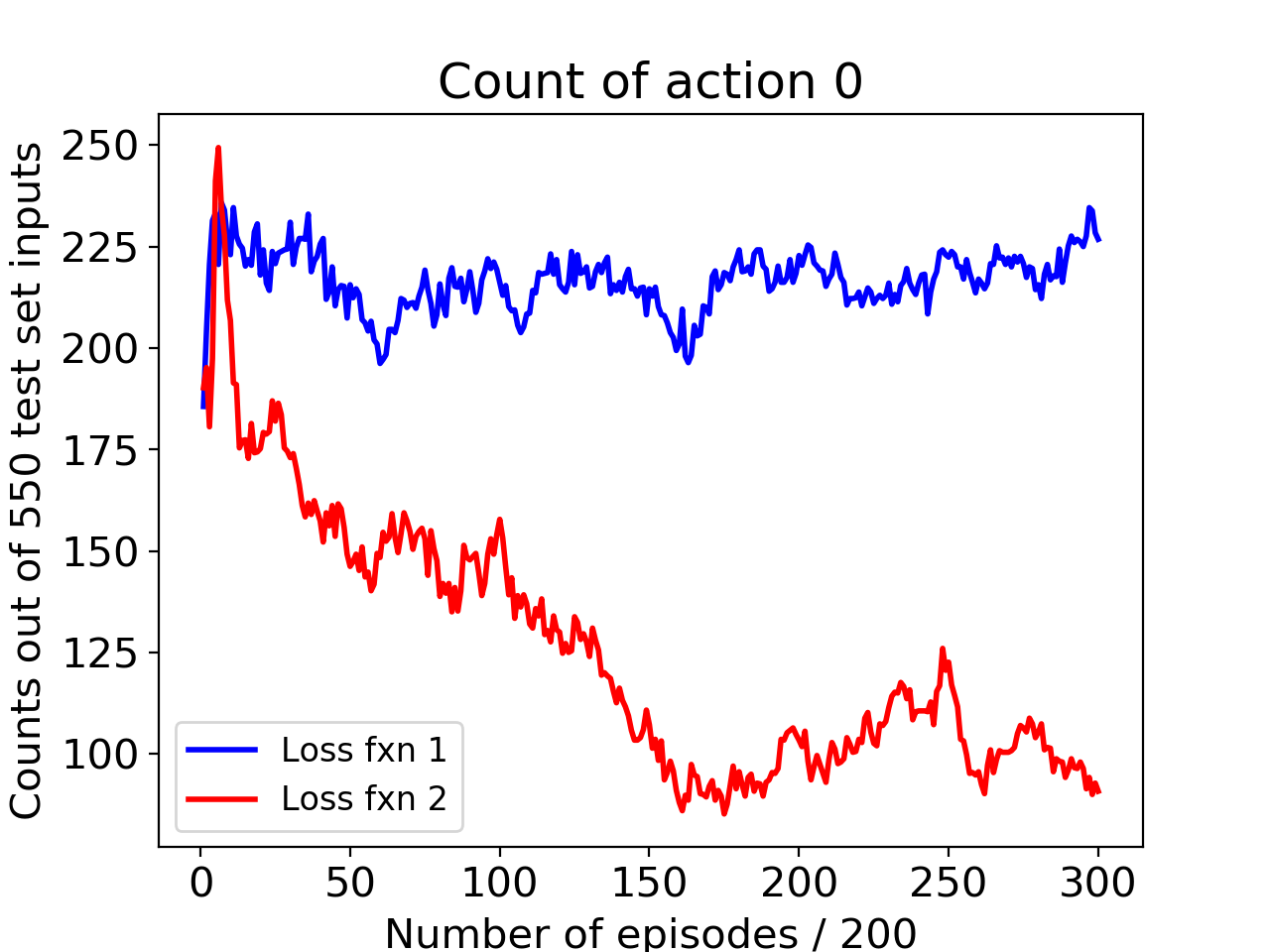}
        \label{fig:count_0}
    \end{subfigure}
    \begin{subfigure}[b]{0.49\linewidth}
        \centering
        \includegraphics[width=\textwidth]{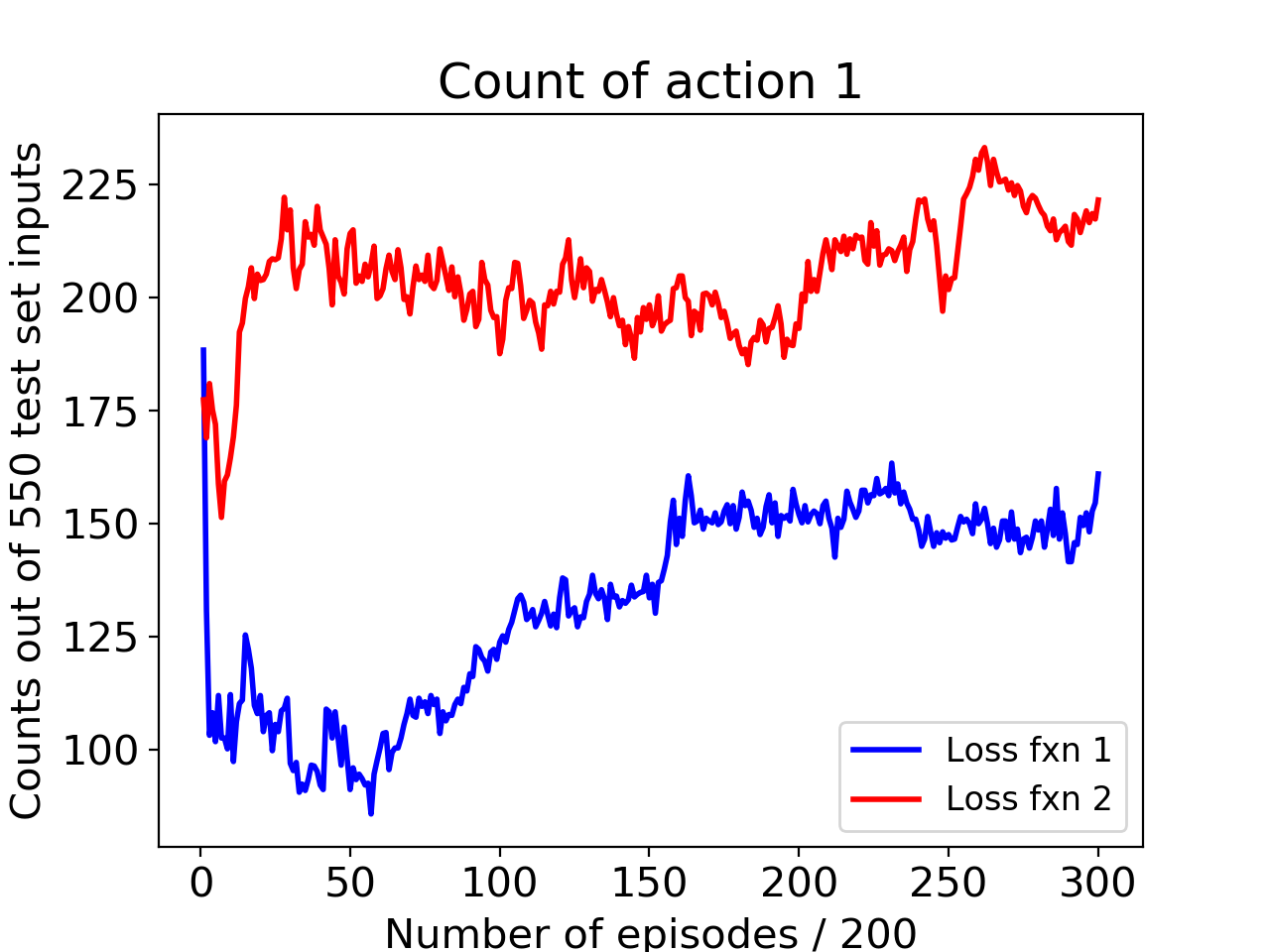}
        \label{fig:count_1}
    \end{subfigure}
    \caption{Action counts in module 2 for per-module CB}
    \label{fig:counts}
\end{figure}

\noindent \textbf{Results:} The dataset of 2689 images is divided into train and test sets of size 2139 and 550 respectively. For training, we use minibatches randomly sampled from the training set and test curves are plotted using the average loss over the complete test set.\footnote{Unlike the synthetic pipeline we do not use average loss over the run of the algorithm here as the number of episodes is much larger than the size of our data set, which means algorithms can overfit to the data set unlike in the synthetic case where we have effectively an infinite data set. So we evaluate a proxy for average loss as the average test performance on held out examples, following standard methodology.} We use the embedding from the penultimate layer of ResNet-50 \citep{he2016deep} as the contextual representation for each image for both adaptive methods. Thus, the context is a 1000 dimensional real valued vector. For per-module CB, the first module's policy network gets the embedding as input whereas the second one gets additional concatenated values of number of faces detected by module 1 and its latency. All networks here have a hidden layer with 256 units with ReLU activations. For evaluating the final loss function of the pipeline, we consider a combination of three metrics:

\noindent \textbf{Pure latency:} Squared loss between the pipeline's latency and a threshold $t_0$: $\ell(a) \coloneqq (\text{latency} - t_0)^2$.

\noindent \textbf{Latency and accuracy:} In addition to the squared distance, we now consider the false negative rate (FNR) of the pipeline for the landmarks detected in each image. Since false negative rate is always in $[0,1]$, it is robust to different number of landmarks in different images as well as different number of predicted landmarks by different actions (5, 27 and 87), unlike a direct classification error in landmark prediction. In this case, $\ell(a) \coloneqq (\text{latency} - t_0)^2 + \text{FNR}$.

\noindent \textbf{Latency, accuracy and false detection penalty:} For the face detection module, in many cases there are non-zero false positives. This further increases the number of false positive landmarks for those cases and therefore we add another penalty of the false discovery rate for face detection.

\begin{table}
    \centering
    \begin{tabular}{|c|c|c|c|}
    \hline
        \#Faces & \#Images & Global CB & Per-module CB \\
        \hline
        $\ge$3 & 124 & 11.82\% & 15.23\% \\
        $\ge$4 & 83 & 18.58\% & 22.51\% \\
        $\ge$5 & 63 & 23.04\% & 24.12\% \\
        \hline
    \end{tabular}
    \caption{Performance percentage improvement over static global policy for global contextual bandit and per-module contextual bandit, broken down by the number of true faces in the image. Numbers shown here are for latency and accuracy loss with $t_0=1.3$.}
    \label{tab:face_comparison}
\end{table}

In our experiments, we choose a value of $t_0 = 1.3$ and $t_0 = 1.7$ for all three loss functions for the pipeline. Note that, if one tries to optimize total latency of the pipeline, then the non-adaptive solution of choosing the cheapest action for both modules works well. Therefore, we choose the bell shaped squared loss for latency which reflects the specification of aiming for a target latency. Figure~\ref{fig:faceSDK-results} shows the observed improvement of the adaptive methods over the static global policy for $t_0 = 1.3$ and $t_0=1.7$. Per-module CB and global CB show improvement for all loss functions against the constant assignment baseline found by greedy hill climbing. The numbers in Table~\ref{tab:face_comparison} show the context-dependency of the pipeline. The benefits of algorithms which are able to effectively utilize context (Global CB and Per-module CB) is really highlighted in the parts of the dataset which contain more than 3, 4 or 5 faces. As the number of faces in an image increases, the percentage gain increases as well. The observed gains of approximately 15, 22 and 24 percent in respecting the utility function are arguably significant for sensitive mission-critical applications. Although these two methods are hard to distinguish on average, we think this is due to the small length of the pipeline and the intermediate context for the second module's policy not being very informative. In order to show that the adaptivity to the final loss function influences the chosen actions, we compare the counts of action $0$ and $1$ for the second modules using the first two loss functions. We show these counts for per-module CB. It can be seen from Figure~\ref{fig:counts} that changing the loss function leads to a change in the chosen actions for the test set.

\subsection{DISCUSSION}
We observe that contextual optimization of software pipelines can provide drastic improvements in the average performance of the pipeline for any chosen loss function. Our experiments show that for small pipelines, both global CB and per-module CB can give potential improvement over a constant assignment. However, these experiments should only be considered as a controlled study of the power of contextual optimization and there are additional caveats which we defer for future work:

\noindent \textbf{Computational overhead:} The loss functions considered by us for the pipeline involve a combination of latency and accuracy. In addition to the pipeline's latency, any metareasoning module will add to the cost. In our experiments, the total time for inference and updates is less than 5-7 \texttt{ms} per input which is orders of magnitude less the the pipeline's latency\besa{how many orders of magnitude?}. Moreover, making the pipeline configurable in real-time might induce further communication/data re-configuration costs. We focus on the potential improvements from adaptivity in this paper and leave the engineering constraints for future work.

\noindent \textbf{Non-stationarity during learning:} For the per-module CB algorithm, the input given to each network is ideally the input for the corresponding module. Changing the configuration of these pipelines can vary the distribution of the inputs to these modules drastically and change in one action changes the input for downstream modules. The pipelines in our experiments do not showcase this issue. We ignore this aspect in our current exposition and leave a more involved study to future work.

\section{CONCLUSION}

We presented the use of reinforcement learning to perform real-time control of the configuration of a modular system for maximizing a system's overall utility. We employed contextual bandits and provided them with a holistic representation of a visual scene and with the ability to both sense and control the parameters of each module. We show significant improvement with the use of the metareasoning methodology for both the face detection and synthetic pipelines. \besa{check before submission:Summarize results here.}  Future directions include studies of scaling up the mechanisms we have presented to more general systems of interacting modules and the use of different forms of contextual signals and their analyses, including the use of more flexible neural network inference methods. 

\section*{Acknowledgements}
This work was done while AM was at Microsoft Research. AM acknowledges the concurrent support in part by a grant from the Open Philanthropy Project to the Center for Human-Compatible AI, and in part by NSF grant CAREER IIS-1452099.


\bibliographystyle{apalike}
\bibliography{arxiv}

\end{document}